\documentclass[letterpaper, 10 pt, conference]{ieeeconf}
\IEEEoverridecommandlockouts
\let\labelindent\relax
\usepackage{graphicx} \graphicspath{ {figures/} }
\usepackage{amsmath,amssymb,mathabx}
\usepackage{algorithmic}
\usepackage[ruled]{algorithm2e}
\usepackage{acronym}
\usepackage{enumitem}
\usepackage{booktabs}
\usepackage{hyperref}
\usepackage{balance}
\usepackage{xspace,setspace}
\usepackage[skip=3pt,font=small]{caption}
\usepackage[skip=3pt,font=small]{subcaption}
\usepackage[dvipsnames]{xcolor}
\usepackage[capitalise]{cleveref}
\usepackage{tabularx,colortbl,multirow,array,makecell}
\usepackage{overpic}
\usepackage{cite}
\usepackage{tikz}
\usepackage{xcolor}

\makeatletter
\DeclareRobustCommand\onedot{\futurelet\@let@token\@onedot}
\def\@onedot{\ifx\@let@token.\else.\null\fi\xspace}

\makeatother

\crefname{algocf}{alg.}{algs.}
\Crefname{algocf}{Algorithm}{Algorithms}

\makeatletter
\def\BState{\State\hskip-\ALG@thistlm}
\makeatother

\makeatletter
\renewcommand{\paragraph}{%
  \@startsection{paragraph}{4}%
  {\z@}{0ex \@plus 0ex \@minus 0ex}{-1em}%
  {\hskip\parindent\normalfont\normalsize\bfseries}%
}
\makeatother

\crefname{algocf}{alg.}{algs.}
\Crefname{algocf}{Algorithm}{Algorithms}

\definecolor{gblue}{HTML}{4285F4}
\definecolor{gred}{HTML}{DB4437}

\acrodef{dof}[DoF]{Degree of Freedom}
\acrodef{vkc}[VKC]{Virtual Kinematic Chain}
\acrodef{tamp}[TAMP]{Task and Motion Planning}
\acrodef{pddl}[PDDL]{Planning Domain Definition Language}
\acrodef{rrt}[RRT]{Rapidly-exploring Random Tree}
\acrodef{ompl}[OMPL]{Open Motion Planning Library}
\acrodef{iws}[IWS]{Iterated Width Search}
\acrodef{bfs}[BFS]{Breadth First Search}
\acrodef{ai}[AI]{Artificial Intelligence}
\acrodef{spt}[SPT]{Scene Parse Tree}
\acrodef{com}[CoM]{Center of Mass}
\acrodef{mcmc}[MCMC]{Markov Chain Monte Carlo}
\acrodef{ged}[GED]{Graph Editing Distance}
\acrodef{cg}[$cg$]{\textit{Contact Graph}}
\acrodef{cg+}[$cg^+$]{\textit{Contact Graph}$^+$}
\acrodef{lbbd}[LBBD]{Logic-Based Benders Decomposition}
\acrodef{komo}[KOMO]{K-Order Markov Optimization}
\acrodef{cpp}[CPP]{Coverage Path Planning}
\acrodef{milp}[MILP]{Mixed-Integer Linear Programming}

\title{\LARGE \bf Dual-arm Cooperation for Tabletop Wiping \\with Temporal Object Removal}

\author{Pengqi Zhang$^{1,2,3}$\quad{}Longchang Yu$^{1,2,3}$\quad{}Daohan Li$^{3}$\quad{}Liang Yan$^{1}$\quad{}Hangxin Liu$^{4\dagger}$
\thanks{$^\dagger$ Corresponding author. Email: \tt{hx.liu@pku.edu.cn}.}
\thanks{$^{1}$ School of Automation Science and Electrical Engineering, Beihang University. $^{2}$ Beijing Institute for General Artificial Intelligence (BIGAI). $^{3}$ Delta Intelligence. $^{4}$ State Key Laboratory of General Artificial Intelligence, School of Intelligence Science and Technology, Peking University.}
}

\begin{document}

\maketitle
\thispagestyle{empty}
\pagestyle{empty}

\begin{abstract}
We devise a dual-arm cooperative planning framework for wiping table surface in the presence of scattered objects. Unlike conventional surface-cleaning strategies that only cover free areas and detour around obstacles, our framework explicitly assigns one arm to rearrange objects while the other performs coverage wiping, thereby enabling access to previously obstructed regions. Specifically, we first construct a set of wiping and object-rearrangement action primitives. Based on these primitives, we introduce Geo-MILP, a geometry-aware task planner formulated as \ac{milp}. By integrating \ac{lbbd}, Geo-MILP iteratively optimizes action sequencing and timing to resolve geometric conflicts, promoting efficient cooperation between wiping and object rearrangement. We then employ \ac{komo} to generate smooth, collision-free motions to ensure feasible execution. Simulation and physical experiments demonstrate that the proposed framework not only increases reachable coverage by enabling access to obstructed areas, but also significantly improves time efficiency compared to \ac{tamp} baselines, reducing planning time by up to $9\times$ and execution makespan by approximately $40\%$.
\end{abstract}

\setstretch{0.985}

\section{Introduction}

Wiping a tabletop surface to remove dust, stains, or liquid spills is a common domestic chore, and automating this task with service robots is highly desirable~\cite{kim2019control,megalingam2025cleaning}. Effective wiping requires the robot’s end effector to thoroughly cover the entire planar surface. This could be accomplished by solving a coverage problem~\cite{mayr2023using,veerajagadheswar2022s,kabir2017automated} in robotics, which has been extensively studied in the \ac{cpp} literature~\cite{galceran2013survey,tan2021comprehensive,bormann2018indoor,wang2025hierarchically}. CPP-based methods have achieved great success in applications such as robotic vacuum cleaning~\cite{zhang2024novel,tan2021comprehensive,bormann2018indoor}, surface inspection and polishing~\cite{wen2022uniform,mcgovern2023general,schneyer2023segmentation,tsai2024path}, where obstacles are typically modeled as static occupancy to be avoided.

\begin{figure}[t!]
    \centering
    \begin{subfigure}[b]{0.49\linewidth}
        \centering
        \includegraphics[width=\linewidth]{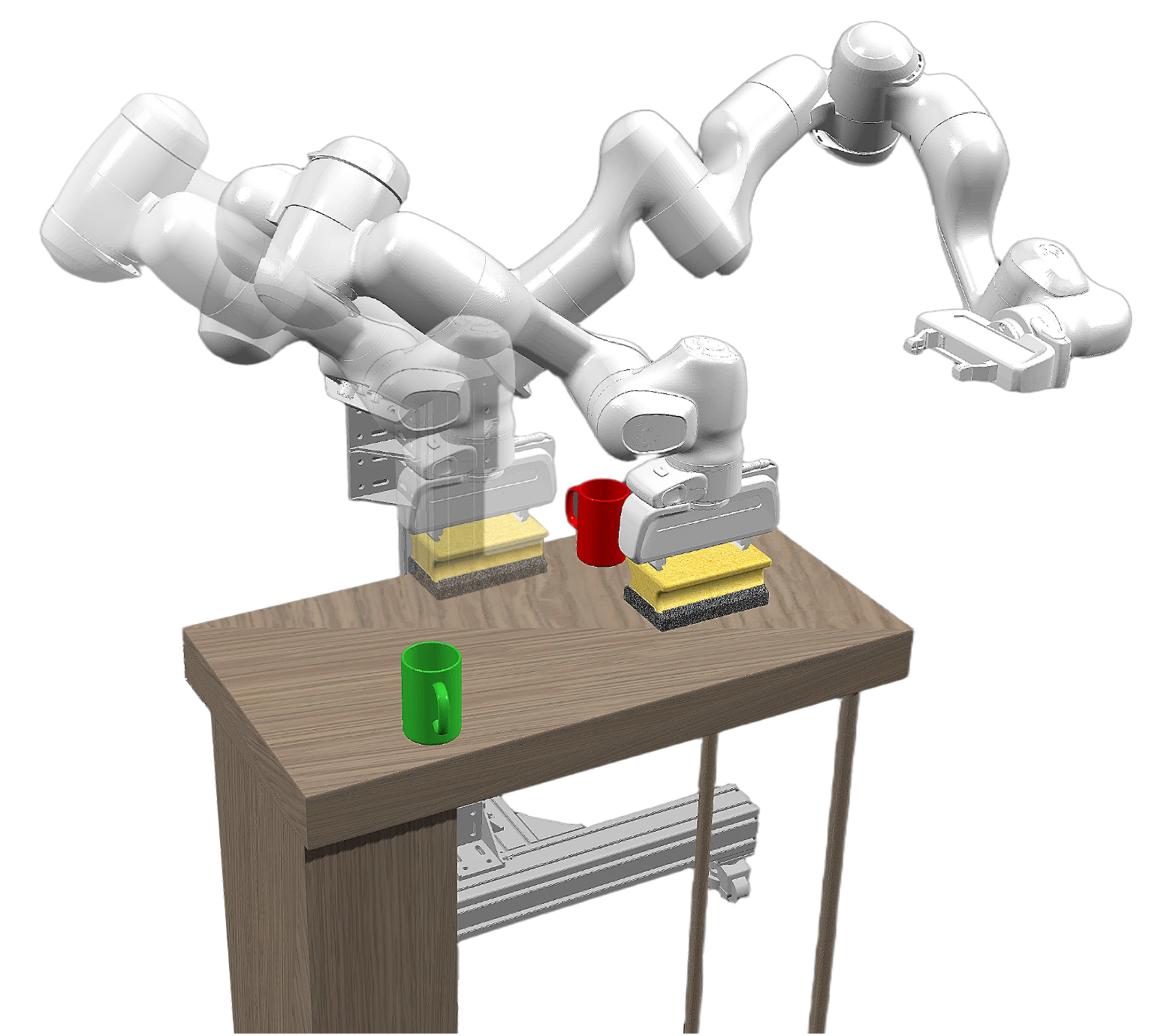}%
        \caption{Classic CPP}
        \label{fig:moti_cpp}
    \end{subfigure}%
    \begin{subfigure}[b]{0.49\linewidth}
    \includegraphics[width=\linewidth]{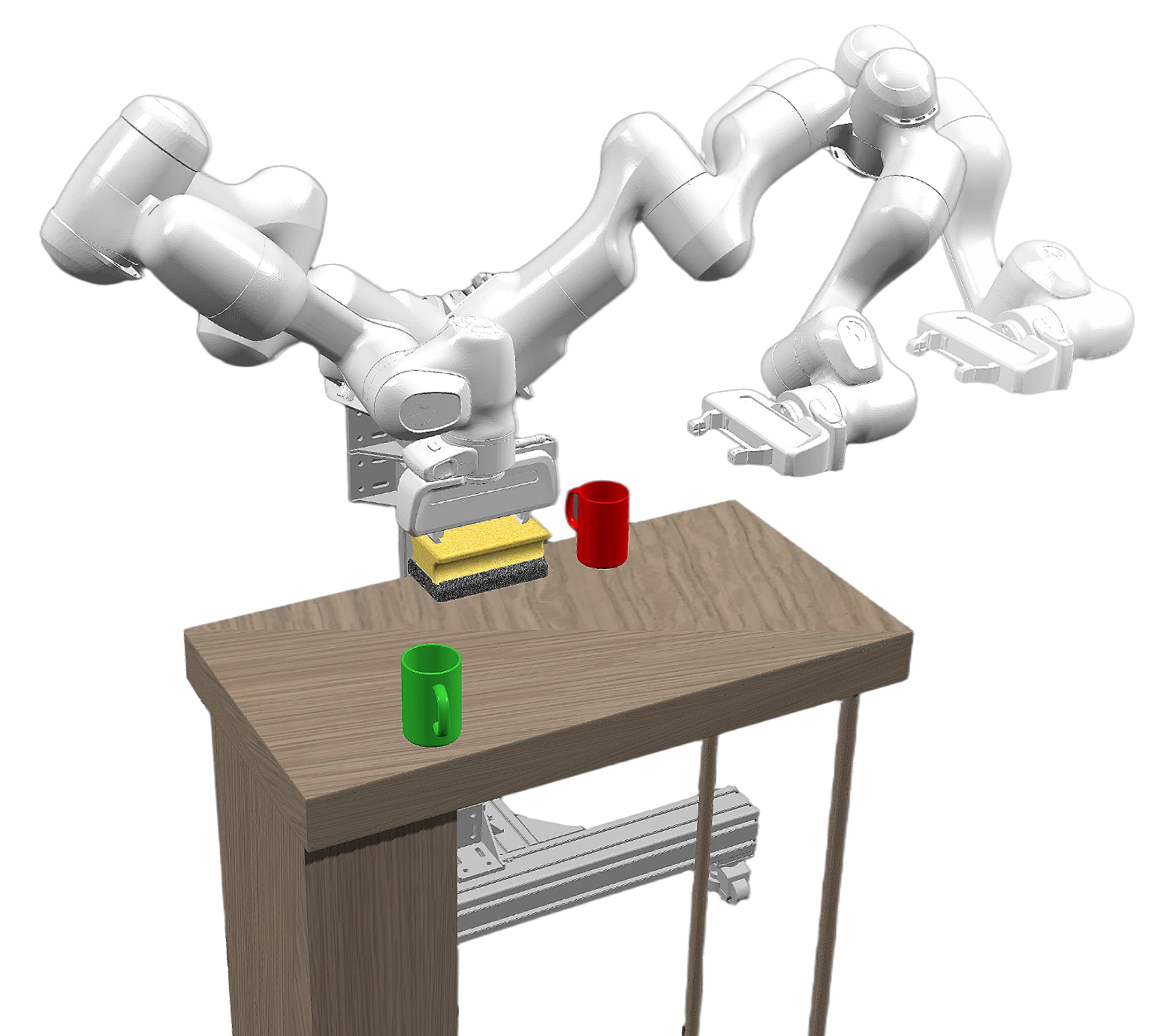}%
        \caption{TAMP}
        \label{fig:moti_tamp}
    \end{subfigure}%
    \\%
    \begin{subfigure}[b]{0.98\linewidth}
        \centering 
        \includegraphics[width=\linewidth]{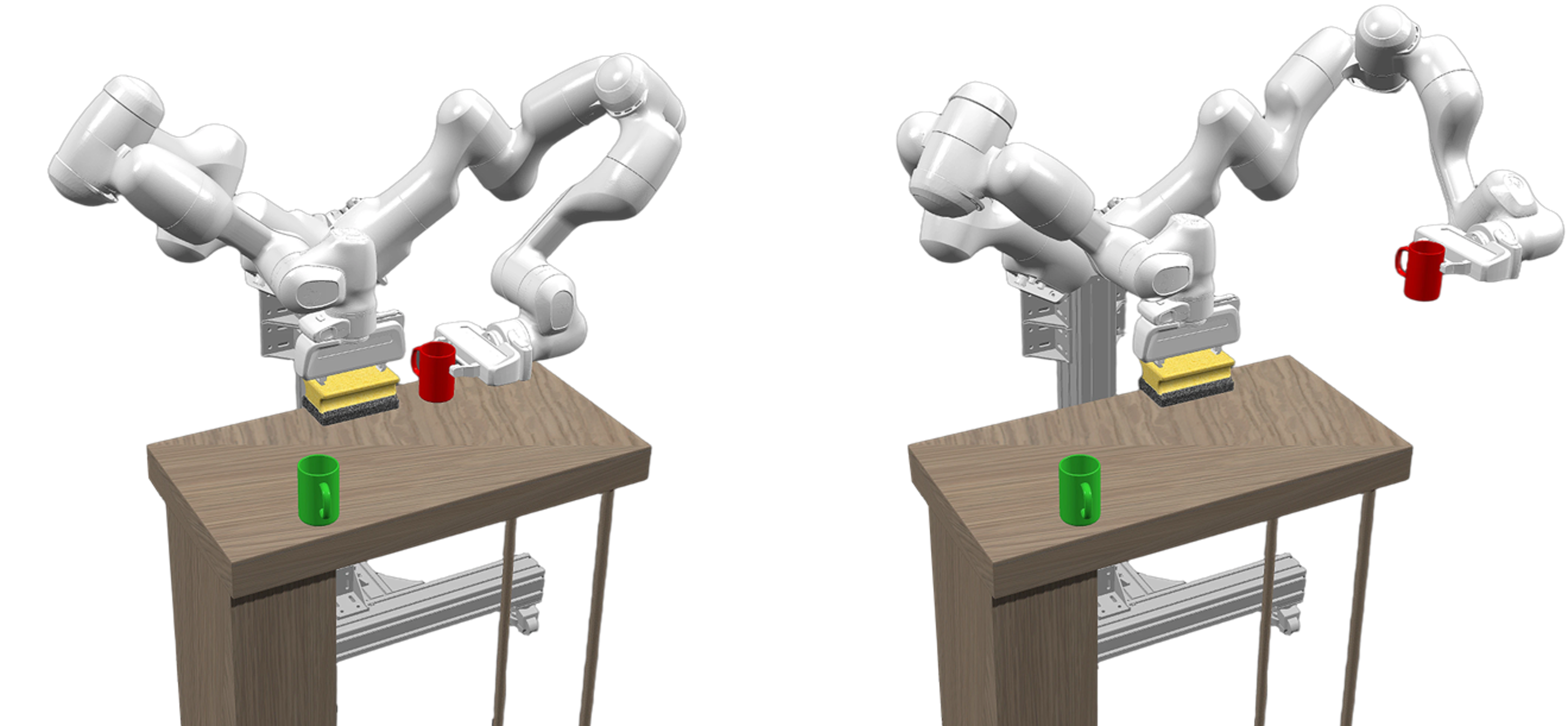}%
        \caption{Ours}
        \label{fig:moti_our}
    \end{subfigure}%
    \caption{\textbf{A tpyical table wiping task.} 
    (a) Classic CPP: The robot detours around obstacles, leaving occluded regions uncleaned. 
    (b) TAMP: The robot strictly alternates between rearrangement and wiping, leading to inefficient execution. 
    (c) Ours (Geo-MILP based TAMP): The planner generates a synchronized schedule where wiping and rearrangement occur concurrently, achieving complete coverage with minimal makespan.}
    \label{fig:motivation}
\end{figure}

However, in real-world domestic settings, desks, kitchen counters, and workbenches are often scattered with everyday objects that partially occlude the tabletop surface. When standard CPP is directly applied in such situation, the robot’s end effector is instructed to detour around objects, leaving large portions of the surface uncleaned (see \cref{fig:moti_cpp}). Moreover, such detouring motions increase the risk of collisions and lead to inefficient trajectories. As a result, achieving complete, safe, and efficient surface coverage in cluttered tabletop environments remains a challenging open problem.

To address this problem, we leverage a dual-arm robot that incorporates temporary object removal to clear obstructions. In particular, one arm performs coverage wiping while the other temporarily removes obstructing objects by picking them up and placing them back after the corresponding surface region has been wiped. This cooperated behavior enables the robot to reach regions that would otherwise be inaccessible. In principle, such behavior can be formulated within a \acf{tamp} framework~\cite{garrett2021integrated,kaelbling2011hierarchical,jiao2021efficient}, where symbolic actions such as pick, place and others are defined at the task level and instantiated into continuous motions after proper sequencing.

However, directly applying conventional TAMP to this problem is often suboptimal for tabletop wiping. Standard TAMP frameworks typically focus on generating feasible sequences of symbolic actions, relying heavily on the assumption of atomic actions and instantaneous state transitions. Under this discrete modeling paradigm, the system state is updated instantaneously between action steps, making it especially difficult to represent the continuous temporal overlap required for coordinated execution. Consequently, without an explicit mechanism to coordinate motion duration, the two arms are forced to operate sequentially rather than concurrently~\cite{mcdermott1998pddl,edelkamp2004pddl2,jiao2021efficient}, leaving one arm idle while the other performs its task, as shown in \cref{fig:moti_tamp}. This not only increases the overall task completion time but also prevents the robot from fully exploiting the advantages of bimanual cooperation. Although TAMP provides a natural formalism for dual-arm tabletop wiping, an additional mechanism is required to coordinately plan bimanual actions to achieve efficient and complete surface coverage.

In this work, we propose a dual-arm cooperative planning framework that explicitly reasons about concurrent wiping and temporary object removal to achieve efficient and complete tabletop cleaning. Our approach first constructs a set of primitive actions for wiping and object handling, and formulates a \acf{milp} problem to optimize their sequencing and temporal coordination, promoting concurrency between the two arms (see \cref{fig:moti_our}). Crucially, to ensure physical feasibility in cluttered environments, we incorporate a \acf{lbbd} strategy that iteratively refines the MILP schedule using geometric feedback from the motion planner. The resulting high-level plan is then refined using \acf{komo} to generate smooth, collision-free motions that are physically feasible for execution on a real robot. The simulation and physical experiment demonstrate that our framework achieves significantly larger surface coverage while reducing task completion time compared to conventional TAMP approaches.

\subsection{Related Work}

\begin{figure*}[t!]
    \centering
    \includegraphics[width=\linewidth]{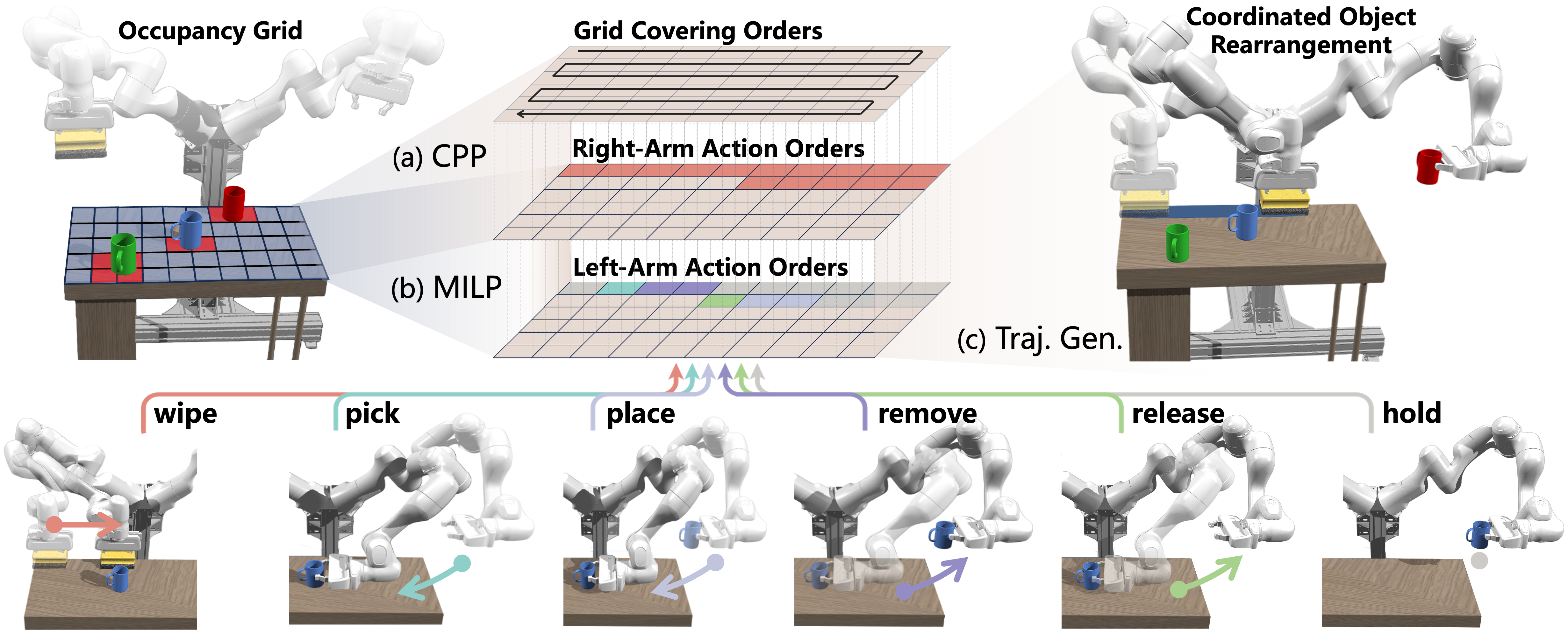}%
    \caption{\textbf{The proposed hierarchical framework.} (a) CPP generates surface coverage waypoints and identifies obstructed nodes. (b) The task planner uses \ac{lbbd} to jointly schedule concurrent bimanual actions. (c) The KOMO-based motion planner translates the discrete schedule into continuous, collision-free joint trajectories.}
    \label{fig:Methods}
\end{figure*}


\textbf{Surface cleaning}
has long been studied as a coverage problem in robotics. Classical \ac{cpp} has succeeded in static surface cleaning tasks like floor and window wiping~\cite{galceran2013survey, tan2021comprehensive}. However, tabletop wiping introduces manipulable clutter. Existing approaches predominantly treat obstacles as immutable constraints, focusing on passive collision avoidance that inevitably leaves occluded regions uncleaned~\cite{bormann2018indoor, martinez2007multi, collins2021scalable, mayr2023using, zhang2024novel}. To guarantee maximal coverage, our formulation integrates CPP with active, reversible object rearrangement to explicitly recover obstructed workspace.


\textbf{Bimanual Manipulation}
research has extensively explored safe motion synchronization~\cite{smith2012dual, kim2019control} and compliance~\cite{mirrazavi2018unified, shi2024safe, jiao2023task}, but it primarily focuses on collision-free execution. Relying solely on low-level coordination limits overall efficiency due to the lack of proactive action scheduling~\cite{pan2021general,li2024synchronized}. To overcome the limitations of purely reactive coordination, our approach elevates dual-arm synchronization to the task planning level. By explicitly optimizing the timing of wiping and rearrangement, we maximize bimanual concurrency rather than merely avoiding collisions.


\textbf{Task and Motion Planning}
seeks to formally integrate discrete high-level reasoning and continuous low-level control~\cite{kaelbling2011hierarchical}. In standard TAMP implementations, the task planning layer typically relies on symbolic planners (e.g., PDDL-based~\cite{mcdermott1998pddl}, VKC-based~\cite{jiao2021efficient}) to resolve causal dependencies~\cite{garrett2020pddlstream,jiao2025integration}. 
However, a fundamental limitation of these approaches is the assumption of atomic actions and instantaneous state transitions. Since they optimize for logical plan length rather than continuous execution time, standard planners lack the intrinsic metric to seek concurrency, typically converging to sequential solutions. To address temporal reasoning, planners such as POPF~\cite{coles2010forward} and LPG~\cite{gerevini2003planning} introduce durative actions and heuristic search over temporal action graphs, enabling partial concurrency. However, they primarily operate in a symbolic domain and do not explicitly incorporate continuous geometric feasibility during planning.
More recent work on multi-arm and makespan-aware planning considers coordination across multiple manipulators~\cite{chen2022cooperative} and explicitly optimizes execution time under coupled task constraints~\cite{hartmann2023towards}, highlighting the importance of temporal coordination in multi-robot systems.
In contrast, our work focuses on a task-specific dual-arm tabletop wiping problem with temporary object removal. We emphasize coordination between coverage wiping and object rearrangement under fixed primitive abstractions, where geometric feasibility is handled by an external motion optimization module.

\subsection{Contribution}

The main contributions of this paper are summarized as follows: (i) We propose a hierarchical dual-arm planning framework that integrates coverage wiping with active, reversible object removal, enabling complete access to occluded regions that are unreachable by conventional collision-avoidance strategies. (ii) We introduce Geo-MILP, a task planner that explicitly optimizes the temporal coordination of dual-arm actions, and combine it with a \ac{lbbd} scheme to iteratively refine the schedule with geometric feedback, ensuring that the generated concurrent plans are not only time-efficient but also physically executable. (iii) We validate the proposed framework which achieves near-complete cleaning coverage in cluttered environments through extensive simulations and physcical experiments.

\subsection{Overview}
The remainder of this paper is organized as follows: \cref{sec:framework} provides a high-level system overview of the proposed coordinated framework and its hierarchical modules. \cref{sec:task_planning} details the formulation of Geo-MILP task planner, and the LBBD-based geometric feedback mechanism. \cref{sec:motion_planning} describes the KOMO-based motion planner used for trajectory generation and feasibility verification. \cref{sec:experiment} presents the simulation and physical experimental results. Finally, \cref{sec:discussion} concludes the paper with discussions.

\section{Coordinated Framework}
\label{sec:framework}

\cref{fig:Methods} provides an overview of our coordinated framework for cluttered tabletop wiping. The key challenge is that maximizing wiping coverage on a tabletop often requires interaction with movable objects, while achieving time efficiency demands temporal coordination between wiping and object rearrangement.  

\subsection{CPP: Tabletop Coverage Path Generation}
We first discretize the tabletop region into occupancy grids and construct a graph over the resulting cells. CPP then computes a coverage-efficient visiting order of waypoints for the wiper to achieve complete surface coverage. At this stage, clutter objects are treated as blocked nodes in the graph: we only identify occluded waypoints but do not generate any interaction or detouring logic. Accordingly, the output of CPP is a coverage-oriented abstraction, including the visiting sequence, adjacency relations, and the subset of occluded waypoints that are infeasible to wipe without manipulation.

\subsection{MILP-based Task Planner: Temporal Scheduling and Dual-Arm Coordination}
Taking the coverage path and blocked-waypoint information produced by CPP as input, we formulate dual-arm coordination as a \ac{milp} problem. The MILP assigns obstacle-handling primitives to the left arm (e.g., \textit{pick}, \textit{place}, and reversible actions that return objects to their original locations) and jointly schedules them on a timeline alongside the right arm's wiping sequence. Its primary goal is to produce a geometrically feasible, time-optimal action schedule that minimizes the makespan under logical precedence and synchronization constraints. To efficiently solve the spatial temporal coupling, we formulate this layer using an optimization-based architecture rooted in \ac{lbbd}. Specifically, the MILP acts as the Master Problem proposing a feasible schedule, while a geometric sub-problem iteratively verifies feasibility and injects logic-based cuts to prune invalid solutions. Through this iterative constraint injection, the task planner refines the schedule until a valid, collision-free task skeleton is found. The solver outputs a time-indexed dual-arm action sequence that balances sequencing and concurrency, avoiding conservative sequential execution.

\subsection{KOMO-based Motion Planner: Trajectory Generation}
The discrete actions scheduled by the MILP is passed to the motion planner. We employ \ac{komo} to translate each scheduled primitive into continuous constraints (e.g., end-effector pose targets for \textit{wipe} and \textit{remove}/\textit{place} conditions) and jointly optimize the resulting trajectories. KOMO handles the robot's kinematic constraints and the scene geometry, enforcing collision avoidance between the arms, the environment, and objects. By solving a trajectory optimization problem, KOMO produces smooth, physically executable, and collision-free trajectories for the scheduled action.

\begin{figure*}[t!]
    \centering
    \includegraphics[width=\linewidth]{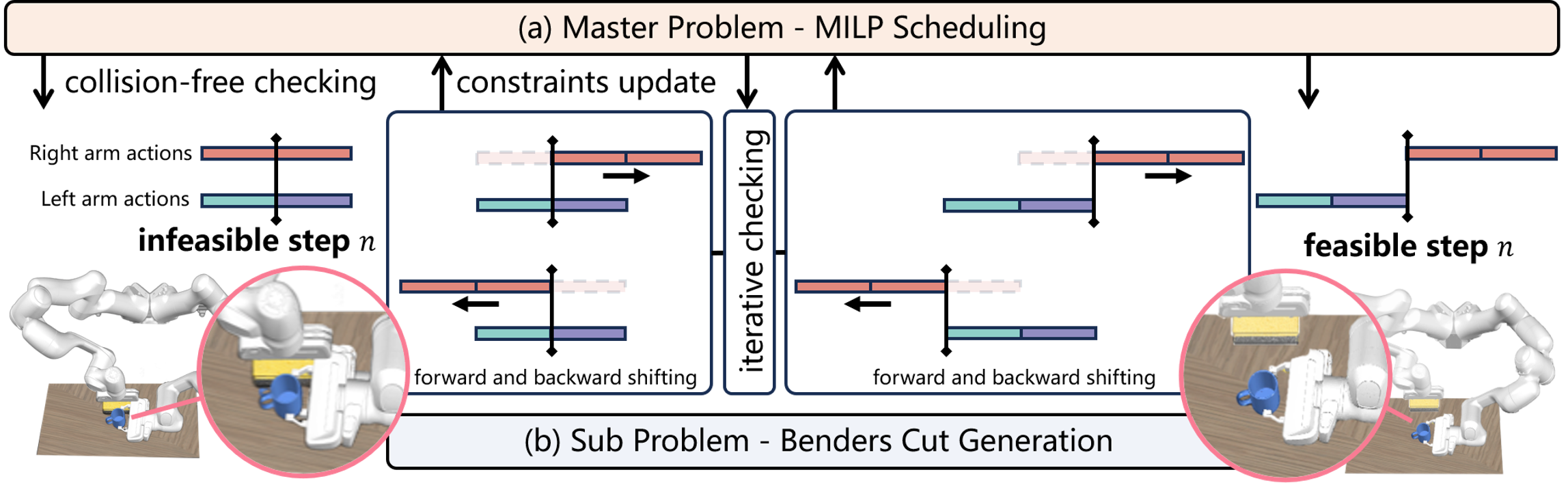}%
    \caption{\textbf{the Geo-MILP Task Planning framework.} 
    The framework employs a \ac{lbbd} scheme to resolve the spatial temporal coupling between the two arms. The Master Problem (MILP) optimizes the high-level task schedule based on logical dependencies, while the Geometric Sub-Problem iteratively verifies the feasibility of candidate schedules and injects logic-based cuts to resolve geometric conflicts. This closed-loop process ensures the generated plan is both time-efficient and physically executable.}
    \label{fig:Geo-MILP}
\end{figure*}

\section{Task Planning via Logic-Based Benders Decomposition}
\label{sec:task_planning}

The primary challenge in bimanual tabletop wiping is the tight spatial temporal coupling between the right arm's continuous coverage and the left arm's discrete rearrangement actions. 
Unlike search-based alternatives, our method leverages \ac{milp} for global coordination, ensuring high search efficiency while maximizing bimanual concurrency.

\subsection{Graph Construction}
\label{subsec:cpp_graph}
As illustrated in Fig. \ref{fig:Methods}, we first abstract the continuous workspace into a graph representation via \ac{cpp}. To transform the surface-covering task into a discrete optimization problem, we perform workspace tessellation. Given a rectangular tabletop $\mathcal{W} \subset \mathbb{R}^2$ and a wiping tool with an effective footprint $\mathcal{F}$, we discretize $\mathcal{W}$ into a grid of cells. We construct a directed graph $\mathcal{G} = (\mathcal{V}, \mathcal{E})$, where each node $v \in \mathcal{V} = \{1, \dots, N\}$ corresponds to a cell center. 

The edge set $\mathcal{E}$ is generated based on a boustrophedon sweeping pattern to ensure full coverage while minimizing redundant traversals. We define an occupancy mapping $\mathcal{O}: K \to 2^{\mathcal{V}}$, where $2^{\mathcal{V}}$ denotes the power set of $\mathcal{V}$. For each obstacle $k \in K = \{1, \dots, M\}$, the mapping $B(k) \in 2^{\mathcal{V}}$ identifies the subset of nodes occluded by the obstacle's physical footprint. Consequently, the coordinated wiping task is formalized as a constrained optimization problem: finding a time-optimal path for the right arm that covers all nodes in $\mathcal{G}$, synchronized with the left arm's rearrangement sequence to clear obstructed regions $B(k)$.

\subsection{Master Problem: Temporal-Logic MILP Formulation}
\label{subsec:milp_formulation}
As shown in Fig. \ref{fig:Geo-MILP}, the Master Problem optimizes the symbolic task sequence and the global timeline, ensuring that all logical dependencies and temporal constraints are satisfied before geometric verification. We formulate this as a MILP model to minimize the total completion time (makespan).


\textbf{1) Continuous Variables ($\in \mathbb{R}_{\ge 0}$):} 

    $T_r^{\mathrm{end}}, T_l^{\mathrm{end}}$: Right and left arm completion times.
    
    $T^{\mathrm{end}}$: Overall makespan of the task ($\max \{ T_r^{\mathrm{end}}, T_l^{\mathrm{end}} \}$). 
    
    $A_v$: The time step index when right arm visits node $v$.
    
    $S_{k}^{\alpha}, E_{u}^{\alpha}$: Start and end times for action primitive $\alpha$ ($\in \{\textit{wipe}, \textit{pick}, \textit{remove}, \textit{place}, \textit{release}\}$) on obstacle $k$ or waypoint $u$. 
    
\textbf{2) Binary Variables ($\in \{0, 1\}$):} 

    $x_{u,v}^t$: Indicator for traversing edge $(u,v)$ ($\in \mathcal{E}$) at discrete step $t$.
    
     $y_{k,u}$: Logical switch variable for resolving geometric collisions between left-arm action on obstacle $k$ and right-arm action at waypoint $u$.



\textbf{3) Objective Function:}
\begin{equation}
    \min \quad T^{\mathrm{end}} \; + \; \varepsilon \sum_{t=2}^{T_{max}} \sum_{i \in \mathcal{V}} x^t_{i i}, \quad \varepsilon \in (0,1) \label{eq:objective}
\end{equation}
where $T_{max}$ is the maximum allowable time steps. We minimize the total makespan while penalizing unnecessary right-arm waiting self-loops ($x_{ii}^t$) to prioritize active movement.

\textbf{4) Subject to Constraints:}

\begin{align}
    & A_v = \sum_{t=1}^{T_{max}} t \cdot \left( \sum_{u \in \mathcal{V}} x_{u,v}^t \right), \quad \forall v \in \mathcal{V} \label{eq:arrival_def} \\
    & A_v \ge E_{k}^{\textit{remove}} + 1, \quad \forall k \in K, \forall v \in B(k) \label{eq:clean_before_enter} \\
    & E_{k}^{\textit{pick}} \le S_{k}^{\textit{remove}}\le E_{k}^{\textit{remove}} \notag\\
    &\le S_{k}^{\textit{place}}\le E_{k}^{\textit{place}} \le S_{k}^{\textit{release}}, \quad \forall k \in K \label{eq:left_seq}
\end{align}

The constraints above bridge high-level task logic with low-level temporal execution. Specifically, Eq. \eqref{eq:arrival_def} retrieves the arrival time $A_v$ from the time-indexed binary decision variables. Eq. \eqref{eq:clean_before_enter} acts as a safety gate, ensuring that the right arm arrives at an occluded waypoint $v$ strictly after the left arm has completed the \textit{remove} action (the $+1$ term enforces the strict inequality $t > t_{remove}$ implied by the discrete state update). Finally, Eq. \eqref{eq:left_seq} enforces a strict causal sequence for the left-arm manipulation primitives, ensuring logically sound obstacle rearrangement.

\subsection{Geometric Sub Problem: Benders Cut Generation}
\label{subsec:lbbd_cuts}

The high-level schedule $\mathcal{S}$ produced by the MILP layer may still contain geometric infeasibilities, such as collisions between the left arm's \textit{pick} motion and the right arm's \textit{wipe} motion. 

We invoke a Geometric Verifier to validate the schedule. As detailed in Algorithm \ref{alg:lbbd}, if a collision is detected between Left Action at obstacle $k$ and Right Action at waypoint $u$, we generate a Separation Constraint (No-good Cut). By introducing a binary decision variable $y_{k,u} \in \{0,1\}$, we inject disjunctive Big-M constraints into the Master Problem:

\begin{align}
    S_u &\ge E_k - \mathcal{M}(1 - y_{k,u}) \label{eq:bigm1} \\
    S_k &\ge E_u - \mathcal{M}y_{k,u} \label{eq:bigm2}
\end{align}
where $\mathcal{M}$ is a large constant. $y_{k,u}$ acts as a logical switch that forces the solver to choose a precedence order: either the left arm completes its action before the right arm starts ($y_{k,u}=0$), or vice versa ($y_{k,u}=1$). The MILP solver selects the precedence direction that is optimal for the current master problem under the accumulated cuts. It then shifts the action windows to obtain a conflict-free schedule after verification. For instance, if a geometric collision occurs during a \textit{pick} action, the solver can seamlessly delay the left arm or command the right arm to wait at a safe node, effectively reshaping the continuous feasible region without the need to backtrack through the discrete task skeleton.

We solve the MILP with Gurobi to optimize \eqref{eq:objective}, obtaining the  makespan $T^{\mathrm{end}}$ and its associated discrete action plan. For downstream path planning, we organize the solution as two time-ordered sequences: $\mathcal{A}^R=\{(t,x_{u,v}^t,\nu_t)\}_{t=1}^{T^{end}}$ and $\mathcal{A}^L=\{(t,\alpha_t,k)\}_{t=1}^{T^{end}}$. The overall framework does not guarantee global optimality for the planning problem, since geometric conflicts are resolved through temporal precedence constraints. We therefore describe the resulting solutions as low-makespan schedules rather than globally optimal ones.

\SetKwInput{KwParam}{Params}

\begin{algorithm}[t]
\caption{Geo-MILP Solver}
\label{alg:lbbd}
\SetAlgoLined
\rmfamily 
\footnotesize

\KwIn{Coverage Graph $\mathcal{G}$; Obstacles $K$; Blocked nodes $B(k)$.}
\KwOut{Optimal conflict-free schedule $\mathcal{S}^*$.}
\KwParam{Big-M constant $\mathcal{M}$; Max iterations $I_{max}$.}

\BlankLine
$\mathcal{C}_{geo} \gets \emptyset$ \;
\For{$i \gets 1$ \KwTo $I_{max}$}{
    \tcp{Master Problem}
    $\mathcal{S} \gets \text{SolveMILP}(\mathcal{G}, K, \mathcal{C}_{geo})$ \;
    \If{$\mathcal{S} = \text{Infeasible}$}{
        \Return{\text{Planning Failed}} \;
    }
    \tcp{Sub-problem} 
    $\mathcal{K}_{coll} \gets \text{CheckGeometry}(\mathcal{S})$ \;
    
    \If{$\mathcal{K}_{coll} = \emptyset$}{
        \tcp{Optimum found} 
        \Return{$\mathcal{S}^* \gets \mathcal{S}$} \;
    }
    
    \ForEach{$(A_L, A_R) \in \mathcal{K}_{coll}$}{
        \tcp{Benders Cuts Generation}
        $S^{A_R}_u \ge E^{A_L}_k - \mathcal{M}(1-y_{k,u})$ \;
        $S^{A_L}_k \ge E^{A_R}_u - \mathcal{M}y_{k,u}$ \;
        $\mathcal{C}_{geo} \gets \mathcal{C}_{geo} \cup \{ \text{Cut} \}$ \;
    }
}
\end{algorithm}

\section{Continuous Motion Generation via Path Optimization}
\label{sec:motion_planning}

After decomposing the overall tabletop wiping task into subtasks and determining the optimal discrete schedule $\mathcal{A}^R$ and $\mathcal{A}^L$, the motion planner must generate continuous, collision-free, and dynamically smooth trajectories. In this section, we utilize the \ac{komo} framework to transform high-level subtask sequences into executable joint configurations by solving a non-linear constrained optimization problem.

\subsection{K-Order Markov Path Optimization}
\label{subsec:komo_fundamentals}

\ac{komo} is an efficient framework for solving trajectory optimization problems~\cite{toussaint2014newton}. It represents a path in the robot configuration space as a discrete sequence of waypoints $\boldsymbol{q}_{0:T}=\{\boldsymbol{q}_0, \dots, \boldsymbol{q}_T\}$, where $\boldsymbol{q}_t \in \mathbb{R}^n$ denotes the joint angles and $n$ represents the manipulator's degrees of freedom (DOF). Under the $k$-order Markov assumption, the cost and constraint functions at any time $t$ are evaluated on $k+1$ consecutive waypoints $\boldsymbol{q}_{t-k:t}$. This allows the optimization problem to be formulated in a general $k$-order non-linear sum-of-squares constrained form:
\begin{alignat}{1}\label{eq:komo_form}
    \min _{\boldsymbol{q}_{0:T}}\quad & \sum_{t=0}^{T} \boldsymbol{f}_{t}\left(\boldsymbol{q}_{t-k: t}\right)^\top \boldsymbol{f}_{t}\left(\boldsymbol{q}_{t-k: t}\right) \\
    \text {s.t.} \quad \forall t: \quad  & \boldsymbol{g}_{t}\left(\boldsymbol{q}_{t-k: t}\right) \leq 0, \quad \boldsymbol{h}_{t}\left(\boldsymbol{q}_{t-k: t}\right)=0 \nonumber
\end{alignat}
where $\boldsymbol{f}_t$ represents the cost terms, while $\boldsymbol{g}_t$ and $\boldsymbol{h}_t$ denote inequality and equality constraints, respectively. Problem \eqref{eq:komo_form} is solved by the Augmented Lagrangian method, which reformulates the constrained problem into an unconstrained one by introducing Lagrange multipliers and penalty terms, making it suitable for handling complex non-linear geometric constraints.

\subsection{Mapping Task Sequences to Optimization Constraints}
\label{subsec:action_mapping}

The core of our motion planner is the automatic translation of the discrete arrival-action sequences $\mathcal{A}^R$ and $\mathcal{A}^L$ into a unified KOMO optimization problem. Each discrete time step $t$ in the MILP schedule corresponds to a fixed time interval in KOMO. For each subtask $(\alpha_t, x_{u,v}^t)$ in the sequences, we impose task-specific constraints over its duration. For instance, during a \textit{wipe} action, the end-effector is constrained to the tabletop surface with a specific orientation, while for the left arm's \textit{pick} and \textit{place} actions, grasping constraints are enforced at specific time-slices.

\subsection{Objective Function: Smoothness and Regularization}
\label{subsec:objective}

The cost function $\boldsymbol{f}$ in Eq. \eqref{eq:komo_form} ensures the generated motion is both natural and efficient.

\textbf{1) Joint Deviation ($k=0$):}
To maintain a favorable configuration and avoid singular poses, we penalize the squared deviation of a configuration $\boldsymbol{q}_{t}$ from a reference configuration $\boldsymbol{q}_{ref}$:
\begin{equation}
    \boldsymbol{f}_{0}(\boldsymbol{q}_{t}) = \sqrt{w_{pos}} (\boldsymbol{q}_{t} - \boldsymbol{q}_{ref})
\end{equation}
where $w_{pos} > 0$ is a scalar weighting factor that regulates the trade-off between task-specific constraints and the preference for a neutral robot posture, providing numerical stability and intuitive tuning of the regularization strength.

\textbf{2) Control Effort ($k=1, 2$):}
We penalize high-order derivatives of the trajectory to ensure smoothness. Specifically, we minimize joint velocities and accelerations through finite-difference operators:
\begin{equation}
    \dot{\boldsymbol{q}}_t \approx \frac{\boldsymbol{q}_t - \boldsymbol{q}_{t-1}}{\Delta t}, \quad \ddot{\boldsymbol{q}}_t \approx \frac{\boldsymbol{q}_{t+1} - 2\boldsymbol{q}_t + \boldsymbol{q}_{t-1}}{\Delta t^2}
\end{equation}
The velocity penalty (first-order difference) penalizes large joint displacements, ensuring the manipulator does not move excessively fast during execution.

\subsection{Geometric and Kinematic Constraints}
\label{subsec:constraints}

\textbf{1) Joint Limits and Collision Avoidance:}
We impose strict inequality constraints to respect the physical joint bounds $\boldsymbol{q}_{\min} \le \boldsymbol{q}_t \le \boldsymbol{q}_{\max}$. Collision-free motion is ensured by evaluating distances $d_{ij}(\boldsymbol{q}_t)$ between all rigid body bounding volumes $K_i$ and $K_j$. We enforce a non-penetration condition with a safety margin $\delta$:
\begin{equation}
    \boldsymbol{g}_{coll}(\boldsymbol{q}_t) = \delta - d_{ij}(\boldsymbol{q}_t) \le 0
\end{equation}
In this work, we use the Gilbert-Johnson-Keerthi (GJK) algorithm to compute Euclidean distances and enforce the accumulated collision measure to be zero.

\textbf{2) Pose Alignment Constraints:}
For precise interaction tasks, such as wiping or grasping, we define position $\boldsymbol{P}(\boldsymbol{q}_t)$ and orientation $\boldsymbol{R}(\boldsymbol{q}_t)$ via forward kinematics. Alignment is enforced via equality constraints:
\begin{align}
    \boldsymbol{h}_{pos} &= \boldsymbol{P}(\boldsymbol{q}_t) - \boldsymbol{p}^\star = 0 \\
    \boldsymbol{h}_{ori} &= \operatorname{Log}(\boldsymbol{r}^\star{}^\top \boldsymbol{R}(\boldsymbol{q}_t)) = 0
\end{align}
where $(\boldsymbol{p}^\star, \boldsymbol{r}^\star)$ is the target pose and $\operatorname{Log}(\cdot)$ produces a minimal 3D orientation error. Additionally, we use axial alignment to constrain specific tool axes $\boldsymbol{u}_i(\boldsymbol{q}_t)$ relative to environmental vectors $\boldsymbol{v}_j$:
\begin{equation}
    \boldsymbol{h}_{axis} = \boldsymbol{u}_i(\boldsymbol{q}_t)^\top \boldsymbol{v}_j - c_i = 0
\end{equation}
where $c_i=1$ enforces parallel alignment.

\section{Simulation and Experiment}\label{sec:experiment}




We evaluate our coordinated dual-arm wiping framework in both simulation and real-robot experiments. The experiments are designed to answer three questions: (Q1) Does explicit object interaction improve achievable coverage beyond obstacle-detouring CPP? (Q2) Does MILP-based temporal coordination improve efficiency over standard TAMP baselines? (Q3) Is the proposed Geo-MILP (LBBD-style) feedback loop necessary to ensure geometric feasibility and reduce expensive geometric re-planning?

\subsection{Experimental Setup}
\begin{figure}[t!]
    \centering
    \includegraphics[width=\linewidth]{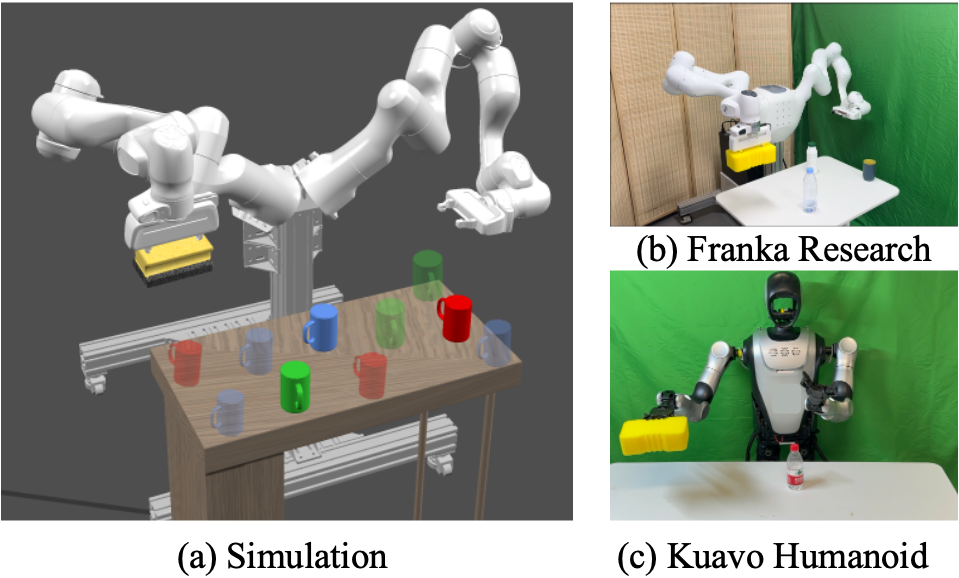}%
    \caption{\textbf{Experimental setups for the dual-arm tabletop wiping task.} (a) The simulation environment used for evaluation. The semi-transparent overlays conceptualize the random spatial variance of the objects. For visual clarity, only a representative subset of the $10$ randomized initializations is shown. (b) Real-world deployment on a dual Franka Research robotic system. (c) Hardware transferability demonstration on the Kuavo humanoid robot platform.}
    \label{fig:Setup}
\end{figure}

The simulated environment features a $0.4\,\text{m} \times 0.75\,\text{m}$ tabletop workspace cluttered with $|\mathcal{K}|\in\{1,2,3\}$ movable objects, as shown in Fig.~\ref{fig:Setup}(a). To perform coverage tasks, the right arm is equipped with a rectangular wiping tool of size $0.09\,\text{m} \times 0.15\,\text{m}$. Each object occludes a specific subset of waypoints (blocked nodes $B(k)$) derived from the wiper's footprint. The dual-arm system is configured with the right arm as the primary wiper and the left arm as the support for object rearrangement. For each difficulty level, we evaluate the performance over $N_{\text{sim}}=10$ randomized configurations to ensure statistical significance.

Real-world validation was conducted on two 7-DoF Franka Research arms in a bimanual configuration. To verify the framework's versatility, we also deploy the algorithm on a Kuavo humanoid robot,  as shown in Fig.~\ref{fig:Setup}(b) and (c). The experimental workspace consists of a $0.6\,\text{m} \times 0.8\,\text{m}$ tabletop, and the physical wiper measures $0.2\,\text{m} \times 0.1\,\text{m}$, adapting to the real-world scale.



\subsection{Behavioral Analysis and Validation}
\label{subsec:qualitative}

We qualitatively evaluate the efficacy of the proposed strategies by analyzing the robot's interaction behaviors in physical environments.

\begin{figure*}[t!]
    \centering
    \includegraphics[width=\linewidth]{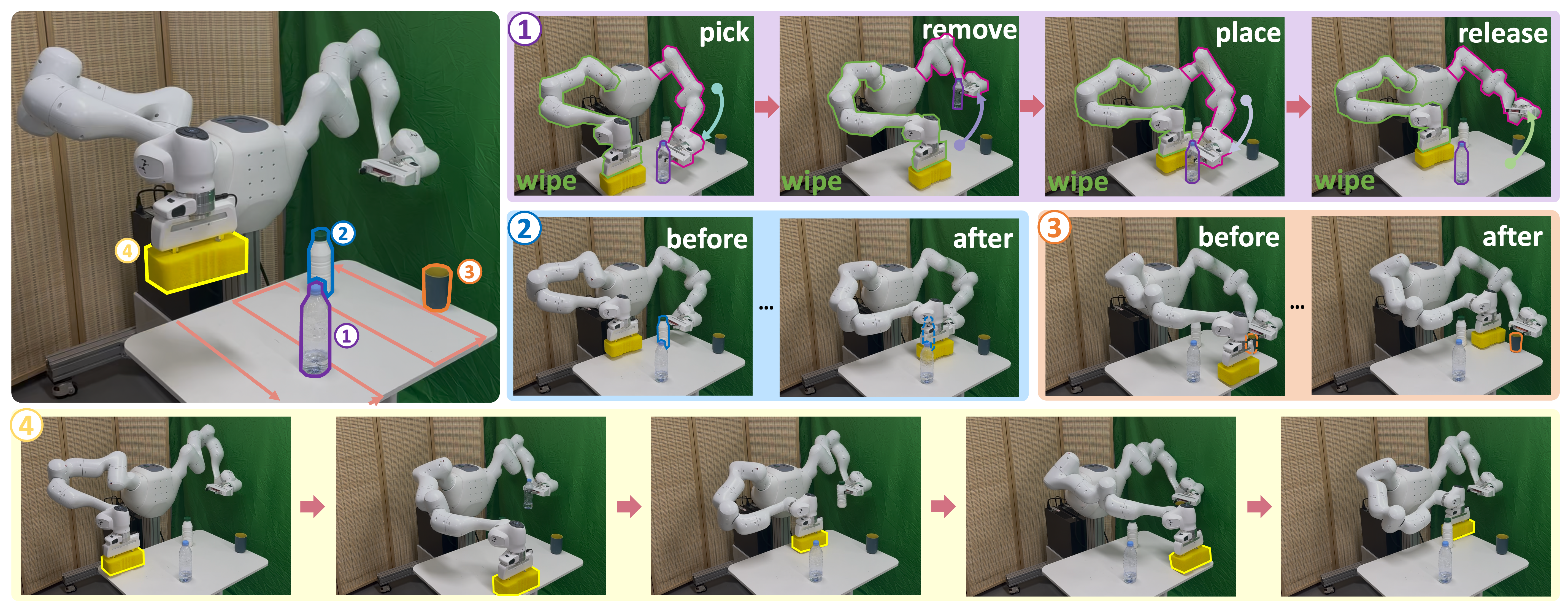}%
    \caption{\textbf{Qualitative execution sequence.} (1) \& (4): Our framework enables true bimanual concurrency: the left arm picks and removes the obstacle, allowing the right arm to wipe the previously occluded surface, followed by placing the object back. (2) \& (3): Comparison of before and after states across different objects. Unlike static single-arm CPP detouring, our reversible relocation strategy achieves complete workspace coverage without leaving uncleaned patches.}
    \label{fig:qualitative results}
\end{figure*}

\textbf{1) Efficacy of dual arm cooperation:}
As shown in the qualitative execution sequence in Fig.~\ref{fig:qualitative results}, Geo-MILP proactively identifies these occlusions and coordinates the support arm to temporarily relocate obstacles.  By mathematically ensuring all unblocked and temporarily unblocked nodes in the CPP graph are visited, our framework inherently guarantees complete coverage of the reachable workspace when the geometric sub-problem yields feasible continuous trajectories, which is visibly confirmed by the qualitative results.

\textbf{2) Physical Execution on Heterogeneous Platforms:}
We further validated the feasibility of the generated plans on two distinct physical platforms. 

On the Franka setup, the system successfully executed coordinated sequences \textit{pick}-\textit{remove}-\textit{wipe}-\textit{place}-\textit{release} for scenarios with up to three obstacles. The motion planner (KOMO) generated smooth trajectories that maintained strict safety margins, even during close-proximity bimanual operations where the wiper passed dangerously close to the support arm. Compared to sequential baselines like PDDLStream, our method exhibited visible concurrency, with the wiping arm moving to the next region immediately while the support arm was restoring the previous object, thereby minimizing idle time. 

The deployment on the Kuavo humanoid demonstrates the framework's kinematic adaptability.  Although the robot's fixed-base configuration limited the reachable workspace to a sub-region of the table, the system successfully performed the coordinated wiping and rearrangement tasks within its range. This validates that our logic-layer scheduling (Stage II) is agnostic to the underlying hardware, provided the motion planning layer (Stage III) is correctly configured for the specific robot kinematics.

\begin{figure}[t!]
    \centering
    \includegraphics[width=\linewidth]{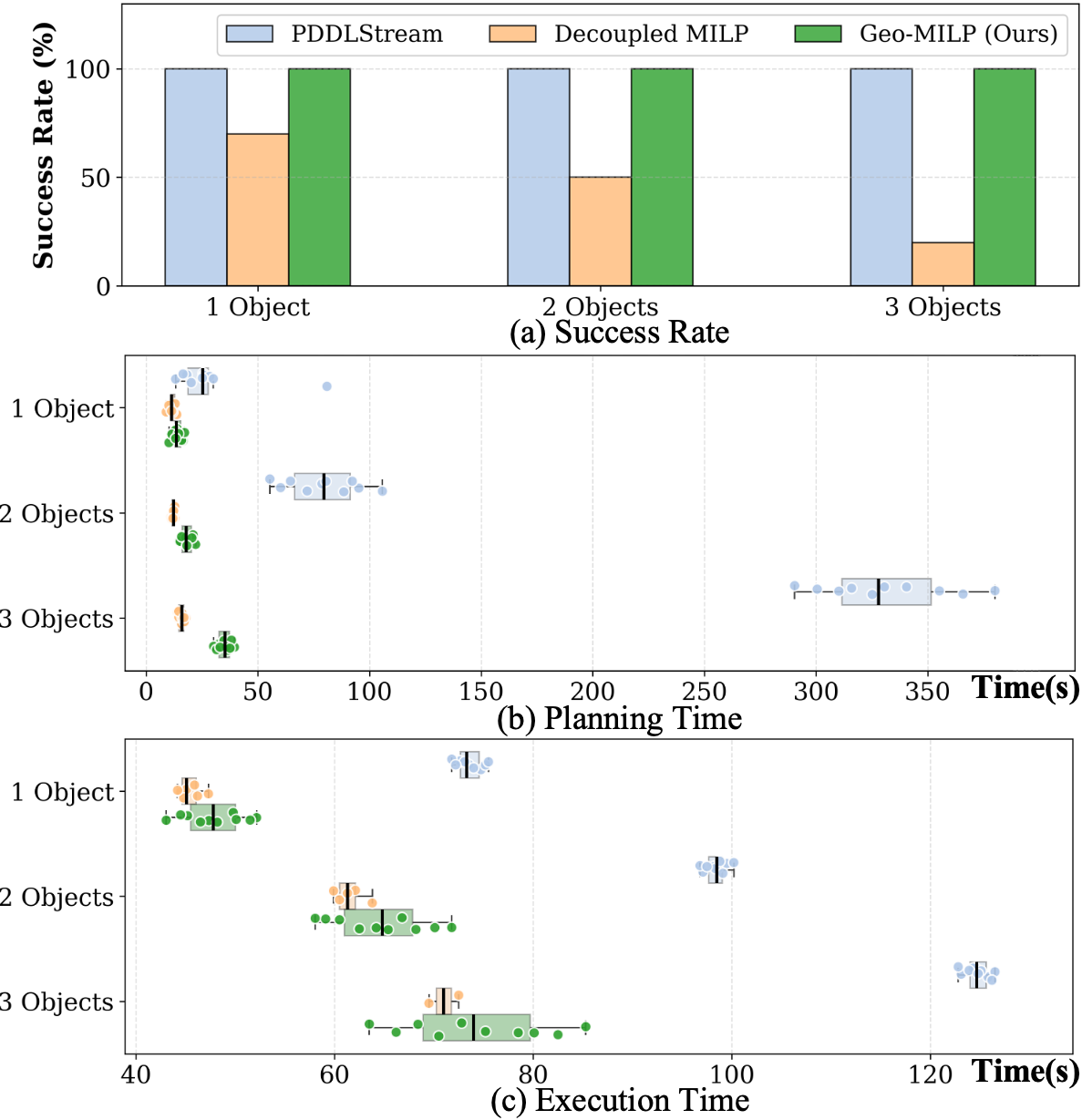}%
    \caption{\textbf{Quantitative performance comparison.} (a) \textbf{Success Rate}: Task success rates across varying clutter complexities. The shared legend applies to all subfigures. (b) \textbf{Planning Time}: Distributions highlighting the tight variance of Geo-MILP compared to search-based baselines. (c) \textbf{Execution Makespan}: Total execution time, showing significant bimanual efficiency gains.}
    \label{fig:quantitative_results}
\end{figure}

\subsection{Efficiency Improvement over PDDLStream}
Fig.~\ref{fig:quantitative_results} compares the performance against standard sequential TAMP (PDDLStream) baselines in terms of both planning and execution efficiency. Since PDDLStream does not explicitly optimize makespan in our implementation, this comparison should be interpreted as a sequential-versus-coordinated comparison rather than a full comparison against all temporal or parallel TAMP methods.

\textbf{1) Planning Efficiency:}
As shown in Fig.~\ref{fig:quantitative_results}(b), PDDLStream suffers from an exponential growth in computation time as the number of objects increases, due to the combinatorial explosion in symbolic state search.
In contrast, Geo-MILP achieves a significant speedup (e.g., approximately $9\times$ faster in 3-object scenes), maintaining a tractable polynomial growth in planning time. The box plots further reveal that while PDDLStream exhibits a long-tailed distribution (high variance), Geo-MILP demonstrates significantly tighter interquartile ranges (IQR), indicating superior algorithmic scalability and stability.

\textbf{2) Execution Efficiency:}
In terms of physical execution, our MILP formulation explicitly promotes concurrency between wiping and obstacle handling.
As shown in Fig.~\ref{fig:quantitative_results}(c), we observe a consistent reduction in the median makespan (e.g., approximately $40\%$ reduction) compared to the sequential baseline. This confirms that our framework not only computes solutions faster but also generates higher-quality schedules that minimize robot idle time.

\subsection{Ablation on Geometric Feedback with LBBD Cuts}

\begin{table}[b]
\centering
\caption{\textbf{Geo-MILP statistics and planning-time breakdown.} The results are reported as mean $\pm$ standard deviation over 10 randomized trials per complexity level.}
\label{tab:lbbd_stats}
\begin{tabular}{cccc}
\hline
$|\mathcal{K}|$ & Geo iters & MILP time (s) & KOMO time (s) \\
\hline
1 & $2.5 \pm 0.5$ & $2.57 \pm 0.83$ & $11.30 \pm 1.45$ \\
2 & $3.8 \pm 1.0$ & $6.39 \pm 2.28$ & $11.79 \pm 0.60$ \\
3 & $5.3 \pm 1.1$ & $19.56 \pm 1.87$ & $15.43 \pm 1.25$ \\
\hline
\end{tabular}
\end{table}

To validate the necessity of the LBBD feedback loop, we compare against a \textit{Decoupled MILP+KOMO} baseline where geometric cuts are disabled.
As illustrated in the planning time distribution (Fig.~\ref{fig:quantitative_results}(b)), the decoupled baseline achieves the lowest computation time (averaging $11$--$17$s) by bypassing the expensive geometric verification loop. However, this efficiency is deceptive: without geometric feedback, the planner generates aggressively concurrent schedules that frequently fail during downstream motion generation due to close-proximity dual-arm conflicts. Consequently, its success rate drops significantly in cluttered scenes (as shown in Fig.~\ref{fig:quantitative_results}(a)).


In contrast, Geo-MILP incurs a moderate overhead (averaging $14$--$37$\,s total planning time) to rigorously enforce geometric consistency. By iteratively injecting the compact temporal cuts, our method resolves conflicts and converges within $2$--$6$ iterations. Crucially, as detailed in Table~\ref{tab:lbbd_stats}, Geo-MILP exhibits remarkable computational stability. Even in dense 3-object clutter, the geometric iteration count remains tightly bounded ($5.3 \pm 1.1$), and the MILP solve time maintains low variance ($19.56 \pm 1.87$\,s). This confirms that our LBBD formulation effectively constrains the search space, avoiding erratic computational spikes and reliably transforming a theoretical schedule into a physically executable one.

Overall, the results show that (i) reversible object relocation is essential for achieving complete wiping coverage in clutter; (ii) explicit temporal coordination via MILP yields substantial efficiency gains over sequential TAMP; (iii) the Geo-MILP feedback loop is critical for producing executable concurrent schedules with high success rates under geometric constraints.

\section{Conclusion and Discussion}
\label{sec:discussion}



In this paper, we proposed a hierarchical dual-arm planning framework for thorough tabletop wiping in cluttered environments. By leveraging \ac{lbbd}, we successfully decoupled symbolic scheduling from geometric verification using Geo-MILP. Unlike conventional search-based TAMP solvers that prioritize logical satisfiability, our MILP-based planner explicitly optimizes the global makespan, promoting maximal concurrency between the two arms. Experimental results demonstrate that our method achieves near-total surface coverage while significantly reducing execution time compared to sequential baselines.

While our framework provides a robust solution, it resolves geometric conflicts solely through temporal precedence constraints under fixed primitive definitions. The possibility of adapting grasp poses, wiping directions, or object placements when conflicts arise is not explored. Future work will incorporate sampled grasp and placement candidates to enable joint spatial reasoning over multiple manipulation alternatives. Additionally, integrating real-time visual feedback will enable more adaptive cleaning in dynamic and unpredictable domestic environments. Also, improving computational efficiency for larger object sets, denser clutter, and finer coverage grids remains important for future research.

\bibliographystyle{ieeetr}
\bibliography{reference}

\end{document}